\pdfoutput=1

\documentclass[letterpaper, 10 pt, conference]{ieeeconf}
\IEEEoverridecommandlockouts   
\newenvironment{IEEEkeywords}{\begin{keywords}}{\end{keywords}}

\usepackage[utf8]{inputenc}
\usepackage[T1]{fontenc}
\usepackage{amsmath}
\usepackage{amssymb}
\usepackage{booktabs}
\usepackage{array}
\usepackage{graphicx}

\graphicspath{{fig/}}
\usepackage{float}

\usepackage{placeins}
\usepackage{cite}
\usepackage{balance}
\usepackage{textcomp}
\usepackage{hyperref}
\hypersetup{
  hidelinks,
  pdfauthor={Zheng Ma, Xiao Jin, Zhenhua Yu},
  pdftitle={Anticipatory Textile MMG--Strain Sensing for Confirm-Then-Assist Soft Knee Exosuits}
}
\usepackage{tikz}
\usetikzlibrary{arrows.meta, positioning, fit, backgrounds, calc}

\definecolor{blkline}{HTML}{1F3A5F}   
\definecolor{blktext}{HTML}{16202B}   
\definecolor{blkctrl}{HTML}{E8EDF3}   
\tikzset{
  blk/.style={draw=blkline, line width=0.5pt, rounded corners=1.2pt,
              fill=white, inner sep=3pt, align=left,
              font=\footnotesize, text=blktext, minimum height=6mm},
  grp/.style={draw=blkline, line width=0.7pt, rounded corners=1.6pt,
              fill=blkctrl, inner sep=3pt},
  flow/.style={-{Stealth[length=4pt,width=4pt]}, line width=0.7pt, draw=blkline},
  fb/.style={-{Stealth[length=3.5pt,width=3.5pt]}, line width=0.5pt,
             draw=blkline, dash pattern=on 2pt off 2pt},
  lbl/.style={font=\footnotesize\itshape, text=blkline, inner sep=1.5pt},
  albl/.style={font=\footnotesize\itshape, text=blkline, fill=white, inner sep=1.2pt},
  hd/.style={font=\footnotesize\bfseries, text=blkline},
}

\renewenvironment{thebibliography}[1]{%
  \begin{oldthebibliography}{#1}%
    \setlength{\itemsep}{0.8pt plus 0.4pt minus 0.2pt}%
    \setlength{\parskip}{0pt}%
}{%
  \end{oldthebibliography}%
}

\title{Anticipatory Textile MMG--Strain Sensing for Confirm-Then-Assist Soft Knee Exosuits}

\author{Zheng Ma$^{1}$, Xiao Jin$^{2}$, and Zhenhua Yu$^{1,*}$%
\thanks{$^{1}$Z. Ma and Z. Yu are with the Department of Computing Science, University of Aberdeen, Aberdeen AB24 3UE, U.K. (e-mail: z.ma.25@abdn.ac.uk; zhenhua.yu@abdn.ac.uk).}%
\thanks{$^{2}$X. Jin is with the Dyson School of Design Engineering, Imperial College London, London, U.K. (e-mail: xj521@ic.ac.uk).}%
\thanks{$^{*}$Corresponding author: Zhenhua Yu.}}

\begin{document}

\maketitle

\begin{abstract}
Soft wearable robots must commit to assistance around movement onset, when the available signals are ambiguous. Surface electromyography (sEMG) detects neuromuscular drive early but degrades under sweat and electrode shift, and kinematic sensors respond only once the limb moves. This letter presents a knitted textile interface that co-samples a three-channel mechanomyography (MMG) array and an anterior strain sensor on a single synchronous $500$\,Hz clock. Across five participants, MMG activation preceded strain-detected knee motion during sit-to-stand by per-channel medians of $322$--$344$\,ms. Sampled on the same clock, the sign of the strain slope resolved movement direction on $649$ of $650$ completed transitions with no training or per-participant calibration, raising four-class balanced accuracy from $90.3\%$ to $95.6\%$. The fused signals drove a two-stage confirm-then-assist policy for a soft pneumatic knee. Across $240$ trials in four participants, the onboard controller entered the assist state on $78$ of $80$ sit-to-stand trials and on none of the $80$ isometric co-contractions. It also entered the assist state on $27$ of $80$ stand-to-sit trials and left it at the hold timeout, motivating an antagonistic flexor gauge to resolve this direction at onset. Co-sampling muscle acoustics with textile strain on one clock lets the inflation command be issued around movement onset.
\end{abstract}


\section{Introduction}

Rising from a chair is among the everyday transitions that become difficult with age~\cite{who2025ageing, hussain2026chairrise}. Assisting such a movement means acting near its onset, before the limb has visibly moved and before its direction can be observed. Soft wearable robots deliver active assistance~\cite{thalman2020review}; exosuits and exoskeletons have lowered the metabolic cost of walking and running~\cite{kim2019reducing, collins2015reducing}; and textile pneumatic systems have been developed for assistive robotics~\cite{shveda2022wearable}. They help only if they capture movement intent reliably and early. How early assistance can arrive is set by both sensing latency and actuation bandwidth~\cite{tiboni2022sensors, nuckols2021individualization, missiroli2020relationship}.

Most wearable sensing architectures rely on kinematic sensors such as inertial measurement units (IMUs), encoders, or foot switches~\cite{mohebbi2020human, torricelli2020benchmarking, tucker2015control, asbeck2015biologically}. These modalities work for cyclic gait, but they register movement only after observable limb displacement has begun. Assistance is then constrained to a reactive response. Surface electromyography (sEMG) captures early neuromuscular drive and has been used to control lower-limb exoskeletons~\cite{ma2019real, gordleeva2020real}. Its wearable use remains limited by sweat-induced impedance collapse~\cite{chi2010dry}, motion artifacts~\cite{de2010filtering, merletti2004surface}, and sensitivity to electrode placement~\cite{beck2008electrode, hermens2000seniam}, which remain open challenges for myoelectric exoskeleton control~\cite{sul2025emg, Toro-Ossaba2025Myoelectric}.

\begin{figure}[!t]
  \centering
  \includegraphics[width=\columnwidth]{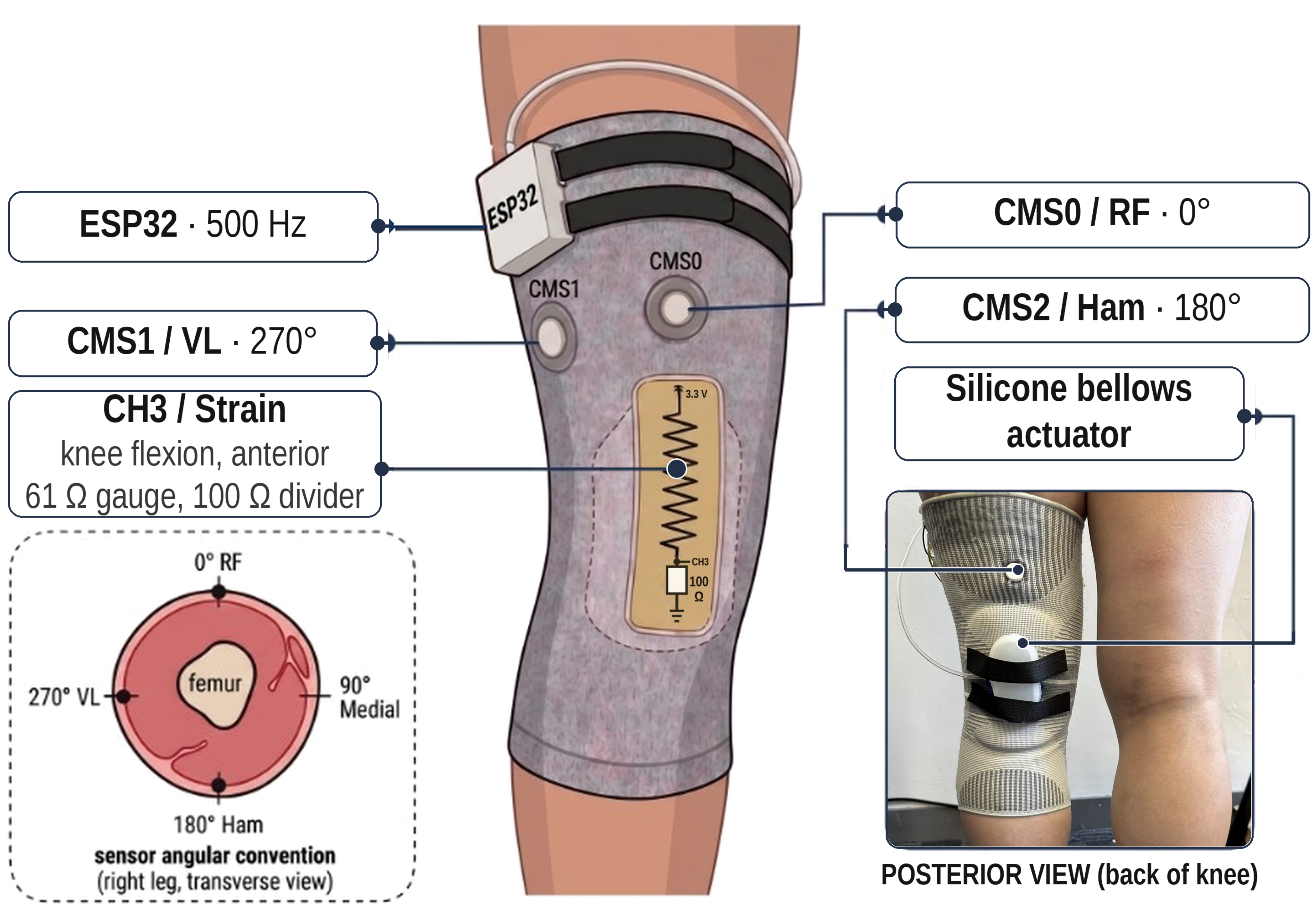}
  \caption{Integrated wearable knee sleeve and dual-modality sensing interface. The anterior illustration locates the ESP32 MCU ($500$\,Hz), the three-channel CMS--MMG array (CMS0/RF at $0^\circ$ anterior, CMS1/VL at $270^\circ$ lateral, CMS2/Ham at $180^\circ$ posterior) and the patellar strain gauge (CH\,3, baseline $61\,\Omega$ on a synchronous $500$\,Hz clock); the transverse key gives the angular convention (right leg). The anterior view and the transverse key are schematics; the posterior inset is a photograph of the fabricated device worn on the leg, in which CMS2 (hamstring), the pneumatic supply tube and the silicone bellows actuator with its retention straps are visible on the back of the knee.}
  \label{fig:prototype}
\end{figure}

In contrast, mechanomyography (MMG) captures low-frequency transverse muscle vibrations through mechanical rather than electrical skin coupling, which leaves the signal independent of skin impedance~\cite{islam2013mechanomyography, smith2017dynamic, orizio1993muscle}. During lower-limb exoskeleton interaction, MMG predicted joint torque comparably to sEMG in dynamic exercise and slightly less accurately under isometric contraction~\cite{caulcrick2021human}. Two problems remain. First, conventional MMG sensors enclose the microphone in a chamber and housing~\cite{caulcrick2021human}, which can create localized pressure points during wear; in this design the casing is embedded in the knitted sleeve behind a filleted contact dome (Sec.~\ref{subsec:textile}). Second, multi-channel MMG patterns can encode movement intent, but amplitude alone does not yield a direction sign without prior training. Concentric sit-to-stand and eccentric stand-to-sit both raise the extensor envelope, and isometric co-contractions vibrate without joint displacement. An untrained magnitude-only scheme therefore cannot resolve assistance direction at onset. MMG with capacitive microphone sensors (CMS) can anticipate IMU-referenced kinematics and tolerate perspiration and attachment variation, gating assistance by a threshold trigger followed by learned torque refinement~\cite{jin2026human}. The amplitude a co-located MMG array reports carries no direction sign on its own. Confirming movement \emph{direction} without training or a rigid external kinematic reference therefore remains open for a soft, low-profile interface.

This letter presents a soft textile sensing and control interface that pairs a three-channel CMS--MMG array with an anterior patellar strain sensor on a synchronous $500$\,Hz clock (Fig.~\ref{fig:prototype}, Fig.~\ref{fig:system}). The MMG array captures anticipatory muscle activation prior to joint motion, while the synchronous strain channel provides a kinematic reference. The main contributions are:

1)~\textbf{Training-free direction confirmation from co-sampled strain}: The sign of the co-sampled strain slope resolves completed transitions with no per-participant calibration ($649/650$ events, against $525/650$ for a supervised MMG classifier), raising four-class balanced accuracy from $90.3\%$ to $95.6\%$ under leave-one-subject-out (LOSO) cross-validation.

2)~\textbf{Textile-integrated co-sampled front end}: A three-channel MMG array and a textile strain gauge share one knitted sleeve and one $500$\,Hz clock, providing the on-sleeve kinematic reference for the $322$--$344$\,ms activation lead without a separate rigid sensor.

3)~\textbf{Confirm-then-assist control}: A feed-forward trigger is combined with an online kinematic gate for a soft pneumatic knee. The onboard controller entered the assist state on $78$ of $80$ sit-to-stand trials and on none of the $80$ co-contractions across four participants ($240$ trials), while stand-to-sit direction remained unresolved at onset, motivating an antagonistic flexor gauge.

\section{System Design and Integration}
\label{sec:hardware}

\begin{figure}[!t]
  \centering
  \includegraphics[width=\columnwidth]{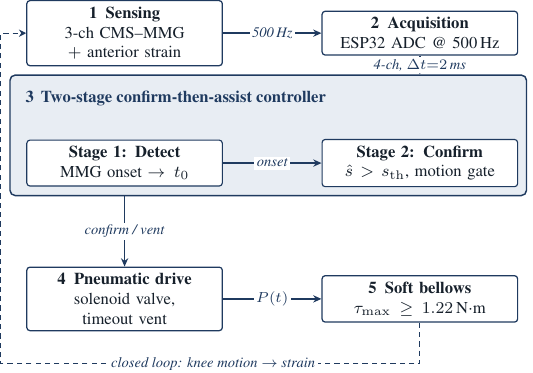}
  \caption{System architecture and signal/control pipeline. Synchronous 500\,Hz sampling of 3-ch MMG and 1-ch strain feeds a two-stage confirm-then-assist policy gating the pneumatic circuit.}
  \label{fig:system}
\end{figure}

A knitted sleeve co-samples a three-channel CMS--MMG array and a co-located textile strain sensor on a single $500$\,Hz ESP32 stream. This stream drives a two-stage confirm-then-assist policy for a soft pneumatic knee actuator (Fig.~\ref{fig:system}).

\subsection{Wearable Textile Design}
\label{subsec:textile}

Wearable MMG requires repeatable sensor--skin acoustic coupling during movement. Sensing is embedded in a tubular knitted sleeve (Fig.~\ref{fig:prototype}) of four layers: elastic textile substrate, three-channel MMG array, anterior patellar strain sensor as knee-flexion proxy, and internal wire harness.

\subsubsection{Sensor Array and Anatomical Selection}
Each MMG unit employs a capacitive microphone sensing architecture~\cite{jin2026human} housed in an axially symmetric casing with a filleted acoustic contact dome to distribute contact pressure. Three CMS units are positioned at a common circumference ($z = 182$\,mm from the distal cuff) over major functional muscle groups (Fig.~\ref{fig:prototype}): Rectus Femoris (RF; \textbf{CH\,0}, anterior), Vastus Lateralis (VL; \textbf{CH\,1}, lateral), and the Hamstring group (Ham; \textbf{CH\,2}, posterior). This ``two agonists plus one antagonist'' configuration spans both extensor and flexor compartments, so an isometric co-contraction registers on both.

\subsubsection{Textile Strain Sensor}
\label{sec:strain}
While the three MMG channels capture muscular \emph{activation}, a fourth channel (CH\,3) tracks joint \emph{kinematics}. A silicone-encapsulated conductive-textile strain sensor with a nominal baseline resistance $R_0 \approx 61\,\Omega$ and relative resistance change $\Delta R/R_0$ reaching $\approx\!3.8\%$ at $25\%$ elongation (Fig.~\ref{fig:strain_e3}) is integrated into the inner anterior lining, spanning the patella, so that it stretches during knee flexion. Wiring is routed internally to prevent snagging. In series with a fixed $100\,\Omega$ resistor on the same patch it forms a resistive divider, sampled by the ESP32 ADC on the same $500$\,Hz clock as the MMG array. Sharing one clock avoids cross-device timestamp alignment. The strain channel serves both subsystems: in sensing it supplies the timing baseline for MMG lead and the slope sign that resolves direction (Sec.~\ref{sec:fusion}); in actuation it gates assistive pressure.

\subsubsection{Acquisition Electronics}
\label{sec:acquisition}
All four channels are digitized single-endedly by the on-board ESP32 successive-approximation ADC at $10$-bit resolution, spanning $0$--$1023$\,counts over the $3.3$\,V rail ($\approx\!3.2$\,mV/count). A fixed $2000$\,\textmu s timer gives a common $500$\,Hz sample instant for the three MMG channels and the strain divider. Samples are timestamped in firmware and streamed over BLE in $15$-sample batches, indexed so that dropped batches are detected rather than silently interpolated. The MMG front end operates at fixed electrical gain throughout, without automatic gain control or per-participant calibration.

\subsection{Soft Inflatable Actuator}
\label{sec:inflatable_exoskeleton}

The assistance mechanism is an inflatable bellows actuator acting across the posterior knee joint. When pressurized during extension, it expands against the posterior thigh and shank to generate an assistive knee extension moment. Sized to the popliteal fossa, the uninflated device measures $70\times40\times16$\,mm with $1.2$\,mm walls and a $64\times34\times6.8$\,mm cavity per half.

The actuator is split-cast from reusable silicone molds featuring an $8$\,mm inlet port and is perimeter-bonded along a $3.0$\,mm rim using uncured silicone ($\approx\!13$\,mL per half). A $0.8$\,mm-thick semi-rigid TPU/PET limiter strip ($70\times30$\,mm) on the outer face blocks outward bulging and redirects expansion force toward the joint. Two low-stretch anchor bands positioned $\pm40$\,mm from the joint center prevent device migration.

\subsection{Pneumatic Circuit}
\label{sec:pneumatic_circuit}

Over a single five-minute static hold, chamber pressure dropped by $2.4\%$ of its initial value. The pneumatic circuit branches to a miniature diaphragm pump and an exhaust solenoid valve through an inline check valve. Inflate, hold, and deflate states are independent. The prototype has a firmware hold timeout, an uncalibrated firmware overpressure trip, and a series hardware emergency-stop button. Its exhaust valve is normally closed, so the chamber does not vent on power loss. A normally-open valve would provide passive venting.

\section{Anticipatory Sensing and Direction Confirmation}
\label{sec:fusion}

We evaluate the dual-modality sensing interface across five participants during daily transitions and test its tolerance to wearable disturbances in single-subject pilots.

\subsection{Strain-Channel Characterization}
\label{sec:exp3_strain}

\begin{figure}[!t]
  \centering
  \includegraphics[width=\columnwidth]{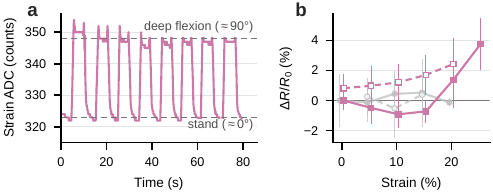}
  \caption{Characterization of the textile knee-flexion strain channel. \textbf{(a)}~In-situ ADC response during continuous sit-to-stand cycles ($500$\,Hz, early divider wiring). \textbf{(b)}~Benchtop tensile calibration as normalized response $\Delta R/R_0$ (mean $\pm$ SD across five cycles) for encapsulated versus unencapsulated sensors under cyclic tensile loading. Squares mark the encapsulated sensor and gray circles the unencapsulated one.}
  \label{fig:strain_e3}
\end{figure}

In situ, cyclic sit-to-stand transitions produced repeatable ADC levels. Values shifted from approximately $323$\,counts during upright standing to $348$\,counts under deep flexion (Fig.~\ref{fig:strain_e3}a). During benchtop tensile loading across $0$--$25\%$ strain with $5$\,mm increments over five cycles (Fig.~\ref{fig:strain_e3}b), the silicone-encapsulated sample had a baseline resistance of $61\,\Omega$, against $31\,\Omega$ for an unencapsulated one. The normalized response is flat within cycle scatter below $15\%$ strain and reaches $+3.8\%$ ($63.4\,\Omega$) at $25\%$, the unloading limb sitting $1$--$2\%$ above the loading limb (Fig.~\ref{fig:strain_e3}b). The sensor shows hysteresis but still separates standing from deep flexion, which is sufficient for onset timing.
While absolute ADC levels reflect the specific divider bias point (median $615$\,counts in the deployed divider configuration, with a $13$-count median excursion), the slope-sign criterion~\eqref{eq:slope} relies strictly on signal derivatives, enabling direct transfer across hardware configurations without threshold recalibration.

\subsection{Experimental Protocol and Cohort}
\label{sec:datasets}

Under a protocol approved by the authors' institutional ethics board, repeated daily lower-limb tasks were conducted in a controlled laboratory, with written informed consent from every participant, including consent to publish video material. Five able-bodied participants (three male, two female; Table~\ref{tab:participants}) spanned heights of $154.0$--$180.0$\,cm and thigh circumferences of $49.0$--$59.8$\,cm across underweight to normal BMI bands, none reporting neuromuscular, musculoskeletal, or mobility impairment. The analysis uses $40$ recordings and $1419$ labeled events: $18$ discrete transition recordings ($650$ transitions, $328$ sit-to-stand and $322$ stand-to-sit) plus $11$ continuous and $11$ stair recordings (stair segments not analyzed). Rest and walking segments in these recordings supply the other $769$ events. All used the same sleeve, placement convention, fixed front-end gain, and $500$\,Hz acquisition chain.

\begin{table}[!t]
\centering
\caption{Physical characteristics of evaluated participants.}
\label{tab:participants}
\vspace{2pt}
\footnotesize
\setlength{\tabcolsep}{4.2pt}
\begin{tabular}{@{}c c c c c c@{}}
\toprule
\textbf{Participant} & \textbf{Age} & \textbf{Sex} & \textbf{Height} & \textbf{Thigh circ.} & \textbf{BMI band} \\
 & (years) & & (cm) & (cm) & \\
\midrule
P01 & 20--30 & M & 170.0 & 55.3 & Normal \\
P02 & 20--30 & M & 180.0 & 59.8 & Underweight \\
P03 & 20--30 & F & 154.0 & 49.0 & Normal \\
P04 & 30--40 & M & 176.0 & 58.5 & Normal \\
P05 & 20--30 & F & 168.0 & 54.0 & Normal \\
\bottomrule
\end{tabular}
\end{table}

The two modalities measure different quantities. The MMG channels report whether and how strongly the thigh muscles contract before joint motion, and the anterior strain channel reports the resulting joint excursion.

\subsection{Signal Processing, Features, and Evaluation Protocol}
\label{sec:methods}

\subsubsection{Conditioning}
Each MMG channel is band-pass filtered with a $4$th-order Butterworth section applied forward and backward (zero-phase, \texttt{filtfilt}) so that no group delay enters a latency measurement. The pass band is $5$--$100$\,Hz for all amplitude, signal-to-noise ratio (SNR), and feature computation, and is narrowed to $10$--$100$\,Hz for onset timing only, where a residual sub-$10$\,Hz sway component otherwise biases the threshold crossing. Band-pass filtering prior to SNR computation prevents low-frequency drift or wetting artifacts from inflating the baseline noise floor. The MMG envelope $e_c[n]$ is the $30$\,ms moving mean of $|x_c[n]|$. The strain channel $s[n]$ is smoothed identically but otherwise unfiltered, preserving its DC level.

\begin{figure}[tb]
  \centering
  \includegraphics[width=\columnwidth]{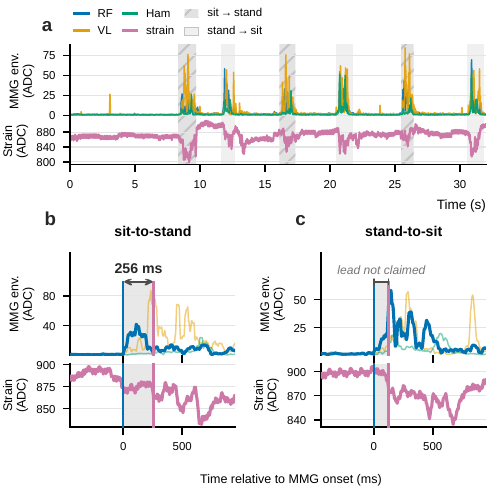}
  \caption{Synchronous 500\,Hz MMG--strain stream (single-subject pilot recording). \textbf{(a)}~Continuous 4-channel stream across transitions. \textbf{(b)},\,\textbf{(c)}~Representative sit-to-stand and stand-to-sit transitions re-zeroed at MMG onset; the bold trace marks the stronger agonist channel. Shaded bands mark the lead of that single event, not the cohort statistics of Fig.~\ref{fig:lead}.}
  \label{fig:cosampled}
\end{figure}

\begin{figure}[tb]
  \centering
  \includegraphics[width=\columnwidth]{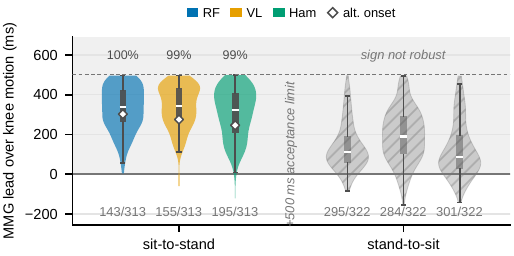}
  \caption{Anticipatory lead of MMG activation over strain-detected knee motion ($n=5$ participants; onset at $3\sigma$ MMG / $5\sigma$ strain). Bottom labels give retained/detected onsets, top labels the positive-lead proportion \emph{among retained} ($\ge 98.7\%$); diamonds ($\diamond$) mark the alternative relative-amplitude criterion (medians $246$--$302$\,ms). The dashed $+500$\,ms limit censors only the \emph{upper} tail of sit-to-stand (slow torso lean before seat-off), so the retained medians are a conservative lower bound. Stand-to-sit (gray) is not claimed anticipatory: leads are mostly positive here, but the sign reverses under the alternative criterion.}
  \label{fig:lead}
\end{figure}

\begin{figure*}[tb]
  \centering
  \includegraphics[width=\textwidth]{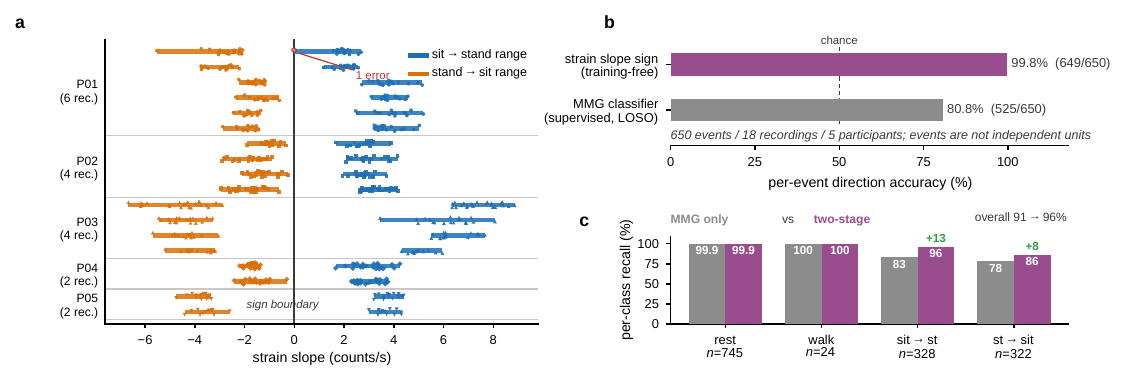}
  \caption{Training-free direction confirmation via strain slope sign ($n=5$ participants, $18$ recordings, $650$ transitions). \textbf{(a)}~Per-recording slope ranges; zero is the criterion, not a fitted threshold. \textbf{(b)}~Per-event comparison against supervised MMG (LOSO, within-event majority vote). \textbf{(c)}~Four-class recall under two-stage fusion; per-class event counts are labeled on the axis.}
  \label{fig:dirpair}
\end{figure*}

\subsubsection{Onset and Anticipatory Lead}
Onsets are referred to a $0.5$\,s pre-transition rest interval of mean $\mu$ and standard deviation $\sigma$. The noise-relative onset is the first index at which the envelope stays above threshold for $H = 15$ consecutive samples, corresponding to a $30$\,ms duration:
\begin{equation}
t_{\mathrm{on}} = \min\bigl\{\, n : e[m] > \mu + k\sigma \ \ \forall\, m \in [n,\, n{+}H) \,\bigr\},
\label{eq:onset}
\end{equation}
with $k = 3$ for MMG and $k = 5$ for the strain deflection in~\eqref{eq:onset}. The stricter strain criterion provides a conservative timing baseline. The lead is $\Delta t_c = (t^{\mathrm{strain}}_{\mathrm{on}} - t^{\mathrm{mmg},c}_{\mathrm{on}})/f_s$, retained when $-200 < \Delta t_c < 500$\,ms. Onsets were also computed under an amplitude-relative criterion at $20\%$ of the peak-to-baseline rise (Sec.~\ref{sec:lead}).

\subsubsection{Signal-to-Noise Ratio}
SNR is defined via~\eqref{eq:snr} as the ratio of band-passed RMS during activity to that of a rest interval,
\begin{equation}
\mathrm{SNR} = 20\log_{10}\!\left(\frac{\mathrm{RMS}_{\text{active}}}{\mathrm{RMS}_{\text{rest}}}\right),
\label{eq:snr}
\end{equation}
and is reported against two denominators. The \emph{local} floor is taken from the resting interval of the same recording; the \emph{common} floor is shared across comparison conditions (dry state for perspiration, centered placement for offsets). Reporting both separates true signal attenuation from baseline noise elevation.

\subsubsection{Strain Slope Criterion}
For an event spanning samples $\mathcal{S}$, the slope is the least-squares fit
\begin{equation}
\hat{s} \;=\; \rho \cdot \arg\min_{a}\ \min_{b} \sum_{n \in \mathcal{S}} \bigl(s[n] - a\,t_n - b\bigr)^2 ,
\label{eq:slope}
\end{equation}
and knee extension is declared when $\hat{s} > 0$, with a fixed threshold of zero and no trained parameters. The polarity constant $\rho \in \{+1,-1\}$ is determined by the resistive divider wiring. Because $\hat{s}$ is a differential slope, the criterion is inherently immune to baseline resistance drift across donnings.

\subsubsection{Windowing, Features, and Cross-Validation}
Recognition uses $0.5$\,s windows at a $0.25$\,s step ($50\%$ overlap). Windows spanning label boundaries are discarded to ensure class purity. Each window yields $7$ features per MMG channel, producing a $21$-dimensional feature space across the array: envelope mean, standard deviation, linear slope, normalized peak position, crest factor, sample-to-sample rate of change, and spectral centroid. Strain contributes two features, excursion range and the polarity-corrected slope of~\eqref{eq:slope}. Absolute strain levels are excluded to prevent baseline drift from biasing classification across donnings.

Classification uses a random forest with $400$ trees and balanced class weights under LOSO cross-validation grouped by participant, with an inference latency of $<0.1$\,ms per window. Window predictions are aggregated within an event by majority vote to match the per-event strain criterion. Events are contiguous runs of ground-truth labels. These metrics are therefore per-event on pre-segmented transitions, and online detection is evaluated in Sec.~\ref{sec:closedloop}. Because rest intervals comprise the majority of data windows, balanced accuracy and macro-F1 are reported alongside overall accuracy. Events from one recording are not independent, so $95\%$ confidence intervals (CIs) were obtained by bootstrapping whole recordings ($10{,}000$ draws).

\subsection{Anticipatory Muscle Activation Lead}
\label{sec:lead}

Muscle activation lead was measured relative to strain-detected knee motion on the shared $500$\,Hz data stream (Fig.~\ref{fig:cosampled}). As shown in Fig.~\ref{fig:lead}, MMG onsets on all three channels preceded anterior strain deflection in five participants. Both onsets were detectable in $313$ of the $328$ sit-to-stand transitions. Within the retention window ($-200 < \Delta t_c < 500$\,ms), median leads ranged from $322$\,ms (Hamstring) to $344$\,ms (VL), with $\ge 98.7\%$ of retained leads positive. Excluded onsets ($38$--$54\%$) primarily reflected prolonged pre-movement postural adjustments (slow torso lean prior to seat-off). Removing the $+500$\,ms retention window yielded cohort medians of $436$--$518$\,ms, and an alternative $20\%$ relative-amplitude criterion gave shorter medians of $246$--$302$\,ms with $\ge 88.7\%$ of leads positive. Stand-to-sit has no such lean and was rarely excluded ($88$--$93\%$ retained). Its lead is not treated as anticipatory, because the extensors contract eccentrically to brake descent rather than to start the movement.

This anticipatory lead is consistent with the mechanical delay between transverse muscle vibration and observable joint rotation. It plausibly combines electromechanical delay ($\sim\!100$\,ms)~\cite{caulcrick2021human} with torso inertial lag and garment compliance. Whereas prior work referenced CMS onset to limb segment orientation via rigid IMUs ($26\pm5$\,ms in~\cite{jin2026human}), these leads are referenced to co-sampled patellar strain, capturing local pre-motion muscle acoustics without external rigid instrumentation. With a proportional valve, this interval could start pressurizing the actuator before seat-off.

\subsection{Direction Confirmation and Recognition}
\label{sec:direction}

Movement direction during postural transfers was evaluated using the strain slope criterion $\hat{s} > 0$~\eqref{eq:slope}. While sit-to-stand (concentric) and stand-to-sit (eccentric) both activate the quadriceps, they produce opposite knee joint rotation. Across $650$ transitions ($18$ recordings from five participants), the zero-boundary rule correctly resolved $649$ events ($99.8\%$). Of the $18$ recordings, $17$ were error-free (Fig.~\ref{fig:dirpair}a). Four of the five participants had zero errors; the only failure occurred in P01 during an initial settling transient ($\hat{s} = -0.0099$\,counts/s). In each of the $17$ error-free recordings, the two transition classes were separated by a margin $>2.0$\,counts/s.
Evaluated per event on the same transitions, a supervised random forest trained on multi-channel MMG features alone reached $80.8\%$ accuracy ($95\%$ CI $69.3$--$90.4\%$) under LOSO cross-validation (Fig.~\ref{fig:dirpair}b). MMG features therefore carry partial direction information. The criterion rests on a physical invariant: knee-angle slope reverses between rising and sitting, so a zero boundary separates the two directions with no trained parameter. A rigid IMU yields the same sign. Knitting the gauge into the sleeve delivers it on the MMG clock, without a housing or a local pressure point.

\textbf{Two-stage recognition.} In the offline pipeline, Stage~1 is the MMG classifier and Stage~2 the strain slope. Four-class recognition was evaluated across all $1419$ recorded events ($745$ rest, $24$ walking, $328$ sit-to-stand, and $322$ stand-to-sit) under LOSO cross-validation. Using MMG alone, Stage~1 recovered rest at $99.9\%$ recall and correctly identified every walking event. The two transition classes were less well separated, with recalls of $83.2\%$ (sit-to-stand) and $78.3\%$ (stand-to-sit). Applying the Stage-2 strain slope to the events flagged by Stage~1 resolved $68$ of the $69$ direction errors, with no rest or walking event reassigned. As shown in Fig.~\ref{fig:dirpair}(c), recall increased to $96.3\%$ for sit-to-stand and $86.0\%$ for stand-to-sit. Overall accuracy increased from $91.1\%$ to $95.9\%$ and balanced accuracy from $90.3\%$ to $95.6\%$ ($95\%$ CI for the gain, $1.3$--$10.4$ percentage points). The residual errors were transitions that Stage~1 missed.

\textbf{Channel ablation.} Table~\ref{tab:ablation} presents leave-one-out and single-channel ablation under the same four-class LOSO protocol. Removing any channel lowered balanced accuracy. VL mattered most: without it, stand-to-sit recall fell from $78\%$ to $54\%$. No single channel matched the full array. Overall, the MMG array carries the amplitude information and the co-sampled strain slope supplies the direction sign it lacks.

\begin{table}[!t]
\centering
\caption{MMG channel ablation. Per-event four-class LOSO (Stage 1, no strain), identical pipeline throughout; $\Delta$ is balanced accuracy relative to the full array.}
\label{tab:ablation}
\vspace{2pt}
\footnotesize
\setlength{\tabcolsep}{4.2pt}
\begin{tabular}{@{}l c c c c@{}}
\toprule
\textbf{Channels} & \textbf{Acc.} & \textbf{Bal.\ acc.} & \textbf{sit-to-stand} & \textbf{stand-to-sit} \\
\midrule
RF\,+\,VL\,+\,Ham (full) & $91.1$ & $\mathbf{90.3}$ & $83$ & $78$ \\
\midrule
$-$\,RF  & $88.4$ & $87.3$ ($-3.0$) & $76$ & $74$ \\
$-$\,VL  & $86.2$ & $84.9$ ($-5.4$) & $86$ & $54$ \\
$-$\,Ham & $90.7$ & $89.9$ ($-0.4$) & $82$ & $78$ \\
\midrule
RF only  & $83.1$ & $81.5$ ($-8.8$) & $82$ & $44$ \\
VL only  & $81.5$ & $79.9$ ($-10.4$) & $59$ & $61$ \\
Ham only & $84.3$ & $82.8$ ($-7.5$) & $80$ & $51$ \\
\bottomrule
\end{tabular}

\vspace{5pt}
{\footnotesize\raggedright
All values in \%. Walking recall is $100\%$ and rest recall is $99.9\%$ across all subsets (omitted for brevity).\par}
\end{table}

\subsection{Validity of the Front End under Interface Disturbance}
\label{sec:validity}

Because the online Stage-1 trigger (Sec.~\ref{sec:twostage}) relies on a fixed amplitude threshold, interface stability was evaluated under two primary wearable disturbances: perspiration and sensor displacement.

\begin{figure}[tb]
  \centering
  \includegraphics[width=\columnwidth]{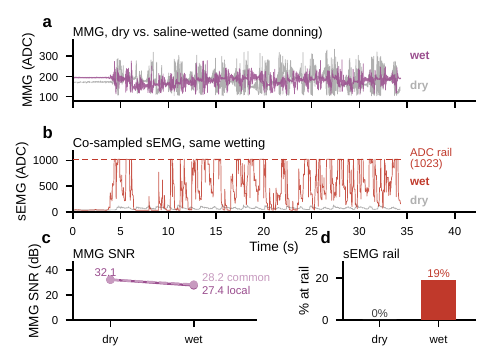}
  \caption{Perspiration robustness of MMG versus co-sampled sEMG within a single donning (single subject). \textbf{(a)}~MMG and \textbf{(b)}~co-sampled sEMG traces, dry versus saline-wetted. \textbf{(c)}~RF-channel MMG SNR during walking against its own resting floor (local) and against the shared dry floor (common). \textbf{(d)}~Fraction of active sEMG samples pinned at the ADC ceiling.}
  \label{fig:sweat}
\end{figure}

\textbf{Perspiration.} We examined interface robustness under perspiration by comparing mechanical and electrical skin coupling under identical wear. As shown in Fig.~\ref{fig:sweat}, saline solution was applied during continuous wear without doffing (preserving placement and pretension), while recording walking with a co-located bipolar sEMG channel ($20$\,mm spacing, Ag/AgCl, fixed-gain differential amplifier into a $10$-bit ADC) synchronously at $500$\,Hz. Neither front end used automatic gain control. After wetting, the RF channel remained continuous and unsaturated, with SNR falling from $32.1$\,dB dry to $27.4$\,dB wetted against its local resting floor ($28.2$\,dB against the common dry floor). In contrast, saline wetting drove sEMG into ADC-rail saturation across $19.1\%$ of active samples. Across three subsequent re-donning sessions, the RF wet-to-dry peak-to-peak ratio stayed within $0.83$--$0.91$, whereas sEMG drifted widely ($0.03$--$3.11$, ranging from severe attenuation to saturation~\cite{chi2010dry}). In this single-subject test, the MMG channel kept a usable signal after wetting without conductive gel or adaptive gain.

\begin{figure}[tb]
    \centering
    \includegraphics[width=\columnwidth]{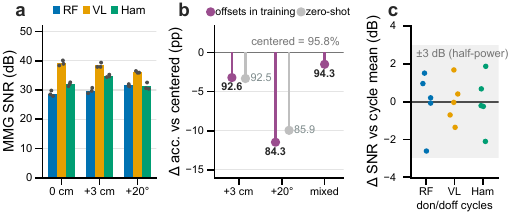}
    \caption{Placement tolerance and re-donning reproducibility (single subject, 3-ch MMG). \textbf{(a)}~Walking SNR across centered, $+3$\,cm proximal, and $20^\circ$ rotational placements, referred to a common noise floor. \textbf{(b)}~Cross-placement classification transfer (Table~\ref{tab:placement}), as accuracy relative to the centered condition: colored markers, the remaining placements in training and the tested one held out (leave-one-placement-out; \emph{mixed} instead pools all three under blocked cross-validation and is not directly comparable); gray markers, trained on centered takes only. \textbf{(c)}~Walking SNR consistency across five don--doff cycles, as deviation about each channel's own mean under a local floor (largest $2.61$\,dB, $n=5$; $\pm3$\,dB half-power reference, not derived from these data); (a) and (c) use different denominators.}
    \label{fig:placement}
\end{figure}

\textbf{Placement.} To evaluate displacement tolerance, three takes each of rest, walking, and sit-to-stand were recorded at nominal centered, $+3$\,cm proximal, and $20^\circ$ circumferential positions, repositioning without re-donning to preserve pretension. Walking SNR stayed within $3.1$\,dB of nominal under a common floor (within $3.4$\,dB local) and above $28.7$\,dB across channels (Fig.~\ref{fig:placement}a). Proximal offset was largely recoverable: with the remaining placements in training it reached $92.6\%$ accuracy against $95.8\%$ centered. Circumferential rotation was not, holding at $84.3\%$ even when the other placements were included in training (Table~\ref{tab:placement} reports every condition with its macro-F1). Across five unassisted don--doff cycles, walking SNR deviated from each channel's mean by at most $2.61$\,dB local ($3.62$\,dB common; worst-case pairwise spread $4.12$\,dB local, $n=5$; Fig.~\ref{fig:placement}c), within a $\pm 3$\,dB half-power band. Sit-to-stand SNR varied more, by up to $10.36$\,dB (RF, local). RF signal RMS during rising exhibited inter-session variability across donnings while resting noise floors remained stable. Such rotational and translational offsets fall outside the electrode-placement tolerance that sEMG protocols assume~\cite{hermens2000seniam, beck2008electrode}.

\begin{table}[!t]
\centering
\caption{Recognition Robustness of the Three-Channel MMG Array Across Controlled Placement Offsets}
\label{tab:placement}
\vspace{2pt}
\footnotesize
\setlength{\tabcolsep}{4.2pt}
\begin{tabular}{@{}l c c c@{}}
\toprule
\textbf{Training set} & \textbf{Centered} & \textbf{$+3$\,cm} & \textbf{$20^\circ$} \\
 & (posC) & (posP3) & (posR20) \\
\midrule
Centered only (zero-shot) & $95.8$ ($0.951$) & $92.5$ ($0.904$) & $85.9$ ($0.838$) \\
All placements pooled     & \multicolumn{3}{c}{$94.3$ ($0.929$)} \\
Leave-one-placement-out  & $90.6$ ($0.887$) & $92.6$ ($0.907$) & $84.3$ ($0.829$) \\
\bottomrule
\end{tabular}

\vspace{4pt}
{\scriptsize\raggedright
Accuracy \% with macro-F1 in parentheses. Three takes per condition within a representative donning session; sliding windows overlap. The leave-one-placement-out rows test a placement absent from training; the pooled row splits blocked window chunks across all three placements, so its value is not directly comparable with them.\par}
\end{table}

The fixed-gain front end tolerated wetting without adjustment. Circumferential rotation caused the largest accuracy loss, so donning alignment should be kept consistent.

\section{Pneumatic Actuation and Control}
\label{sec:actuation}

\subsection{Actuator Saturation Output}

To avoid overestimating worn performance on rigid testbeds, actuator output was characterized directly against the textile sleeve. At full inflation under maximum operating pressure in the textile sleeve configuration, the actuator lifted a proof mass $m = 4.145$\,kg, so that
\begin{equation}
\tau_{\max} \;\ge\; m\,g\,r \;=\; 4.145 \times 9.81 \times 0.030 \;=\; 1.22\,\mathrm{N}\!\cdot\!\mathrm{m},
\label{eq:torque}
\end{equation}
about the nominal popliteal moment arm $r \approx 30$\,mm. Equation~\eqref{eq:torque} is a bench sizing bound lifted against gravity, measured before human evaluation.

\subsection{Two-Stage Control Policy}
\label{sec:twostage}

The bellows actuator is governed by the two-stage confirm-then-assist policy: a feed-forward myographic trigger paired with a kinematic safety gate.

In \emph{Stage~1 (detection)}, onset is flagged when any MMG channel's envelope exceeds eight times its tracked resting level (and $2.5$\,counts), opening the Stage-2 verification window. The $30$\,ms envelope avoids sliding-window buffering delay.

In \emph{Stage~2 (confirm)}, an online kinematic gate verifies movement onset before authorizing sustained assistance. For the divider wiring used here ($\rho = -1$), knee extension yields a negative raw ADC slope. To suppress ADC quantization noise ($\pm1$--$2$\,counts), trailing $300$\,ms linear regression is computed ($N=150$; $\sigma_{\hat{s}}\approx0.9$\,counts/s for $1$-count white noise), and a deadband $s_{\mathrm{th}} = 4.0$\,counts/s ($\approx\!4\sigma_{\hat{s}}$) is applied. The deadband was selected on $70$ actions recorded from P01 in two donnings before, and separate from, the online sessions ($25$ sit-to-stand, $25$ stand-to-sit, $20$ co-contractions). The absence of false confirmation among its $45$ non-extensor trials is a calibration-set figure, not an estimate of generalization.

\subsection{Timing Budget}

The pneumatic step response was measured on the bench over $n=8$ inflation cycles at two nominal pressure set-points, sampled at $500$\,Hz (mean $\pm$ SD): inflation time constant $\tau_{\text{inf}}=412\pm39$\,ms, $90\%$ plateau rise in $844\pm88$\,ms, active deflation in $165\pm19$\,ms. These response times are consistent with a flow-limited miniature diaphragm pump operating without an accumulator.

\begin{figure}[!t]
  \centering
  \includegraphics[width=\columnwidth]{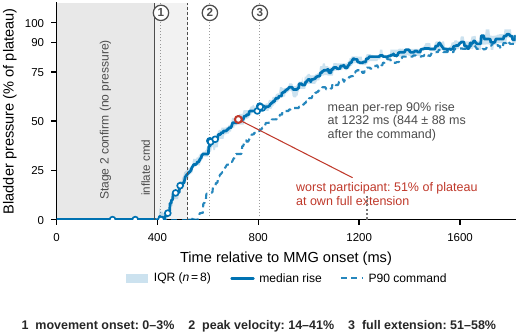}
  \caption{Modelled pneumatic assistance arrival relative to MMG onset under two-stage control. Solid: median step response over $n=8$ benchtop inflation cycles ($\tau_{\text{inf}}=412\pm39$\,ms, mean $\pm$ SD), IQR shaded; dashed: P90 command time ($+520$\,ms). The command time is the on-body Stage-2 latency measured in the four participants evaluated online; the kinematic milestones come from the five sensing participants, so open circles give the modelled pressure at each participant's milestone, obtained from the bench response at the measured confirmation latency, and the footer key the range at each.}
  \label{fig:assist_timeline}
\end{figure}

The chamber was not pressurized on the body, so Fig.~\ref{fig:assist_timeline} models the timeline: the bench step response is applied at the measured confirmation latency, a median $388$\,ms after Stage~1 (Sec.~\ref{sec:closedloop}), and evaluated at kinematic milestones from the strain channel of the five sensing participants. In this model the command precedes strain-detected movement onset in three of the five participants, and pressure reaches $0$--$3\%$ of plateau at onset, $14$--$41\%$ (median $39\%$) at peak extension velocity, and $51$--$58\%$ (median $57\%$) at upright extension. With the $844$\,ms rise time, peak pressure would arrive during early quiet standing, and active exhaust empties the chamber in $165$\,ms before the next flexion.

\subsection{On-Body Online Evaluation}
\label{sec:closedloop}
The integrated two-stage control strategy (Stage-1 trigger $\to$ Stage-2 confirmation $\to$ inflation command) was run in real time on four of the five participants of Table~\ref{tab:participants} (P01--P04), with the pump and pressure transducer disconnected. Each ran $60$ consecutive trials under the onboard $500$\,Hz loop: $20$ sit-to-stand, $20$ stand-to-sit, and $20$ isometric co-contractions, with one firmware build and no per-participant tuning. P05 was also recorded, but the controller board reset during acquisition and that session was excluded in full.

During dynamic transfers, the initial acoustic burst triggered Stage~1, after which the anterior strain sensor was tracked over the $300$\,ms regression window. The controller entered the assist state on $78$ of the $80$ sit-to-stand trials; the Stage-2 gate rejected the other two. Co-contractions reached Stage~1 in $67$ of $80$ trials and the gate rejected all $67$, so none entered the assist state. For stand-to-sit, $27$ of $80$ trials passed the gate and entered the assist state, and none of the $80$ was resolved as flexion online; the assist state was released at the $1.5$\,s hold timeout. Table~\ref{tab:closedloop} gives the per-participant counts. Scored by controller decision, $211$ of $240$ trials behaved as intended.

\begin{table}[tb]
\caption{On-body online decisions, $60$ consecutive trials per participant under the $500$\,Hz onboard loop}
\label{tab:closedloop}
\vspace{2pt}
\centering
\renewcommand{\arraystretch}{1.20}
\footnotesize
\setlength{\tabcolsep}{3.0pt}
\begin{tabular}{@{}l c c c c c@{}}
\toprule
\textbf{Assist decision} & \textbf{P01} & \textbf{P02} & \textbf{P03} & \textbf{P04} & \textbf{Pooled} \\
\scriptsize($20$ trials each) & & & & & \\
\midrule
Sit-to-stand (desired)        & $19/20$ & $19/20$ & $20/20$ & $20/20$ & $78/80$ \\
Stand-to-sit (undesired)      & $6/20$  & $8/20$  & $12/20$ & $1/20$  & $27/80$ \\
Co-contraction (undesired)    & $0/20$  & $0/20$  & $0/20$  & $0/20$  & $0/80$ \\
\midrule
\addlinespace[1pt]
Correct decision                & $53/60$ & $51/60$ & $48/60$ & $59/60$ & $211/240$ \\
\addlinespace[1pt]
\bottomrule
\end{tabular}

\vspace{4pt}
{\scriptsize\raggedright
Scored by controller decision: entering the assist state is correct for sit-to-stand and incorrect otherwise. Co-contractions reached Stage~1 in $67/80$ trials and all $67$ were rejected by the Stage-2 gate.\par}
\end{table}

Stage-2 confirmation followed Stage~1 by a median $388$\,ms (IQR $339$--$464$, range $300$--$732$\,ms, $n=78$), with the $300$\,ms regression window setting the floor. After the trigger, the online gate regresses over the first $300$\,ms of strain, before substantial joint displacement; the $99.8\%$ of Sec.~\ref{sec:direction} was instead computed over completed transitions. Within that window the anterior gauge deflected the same way for extension and flexion, which we attribute to local textile--tissue bunching, so a single gauge certifies that movement has begun but not its direction. An antagonistic flexor gauge could supply the missing sign at onset without a rigid enclosure.

\section{Conclusion}
\label{sec:conclusion}

This letter presented a soft textile interface combining anticipatory MMG sensing with a training-free strain direction criterion for pneumatic knee assistance. The two-stage policy issues its inflation command around movement onset, and chamber pressure then develops through the extension. Across five participants, the strain-slope sign raised four-class balanced accuracy from $90.3\%$ to $95.6\%$ under LOSO cross-validation, and MMG activation preceded strain-detected knee motion by median leads of $322$--$344$\,ms.

Future work will test powered on-body assistance and evaluate in-situ joint torque and metabolic reduction across a broader cohort during extended daily activities. We will also investigate strain tracking across multi-position donnings, incorporate fast proportional valves to deliver assistive pressure prior to seat-off, and integrate an antagonistic flexor gauge to resolve movement direction at onset. The acquisition firmware, analysis scripts, and derived event-level data will be released publicly upon acceptance.

\balance
\bibliographystyle{IEEEtran}
\bibliography{ref}

\begin{thebibliography}{10}
\providecommand{\url}[1]{#1}
\csname url@samestyle\endcsname
\providecommand{\newblock}{\relax}
\providecommand{\bibinfo}[2]{#2}
\providecommand{\BIBentrySTDinterwordspacing}{\spaceskip=0pt\relax}
\providecommand{\BIBentryALTinterwordstretchfactor}{4}
\providecommand{\BIBentryALTinterwordspacing}{\spaceskip=\fontdimen2\font plus
\BIBentryALTinterwordstretchfactor\fontdimen3\font minus
  \fontdimen4\font\relax}
\providecommand{\BIBforeignlanguage}[2]{{%
\expandafter\ifx\csname l@#1\endcsname\relax
\typeout{** WARNING: IEEEtran.bst: No hyphenation pattern has been}%
\typeout{** loaded for the language `#1'. Using the pattern for}%
\typeout{** the default language instead.}%
\else
\language=\csname l@#1\endcsname
\fi
#2}}
\providecommand{\BIBdecl}{\relax}
\BIBdecl

\bibitem{who2025ageing}
{World Health Organization}, ``Ageing and health,'' WHO Fact Sheet, 2025,
  [Online]. Available:
  \url{https://www.who.int/news-room/fact-sheets/detail/ageing-and-health}.

\bibitem{hussain2026chairrise}
M.~A. Hussain, S.~Naheed, K.~Saeed, A.~Karim, F.~Ahmad, R.~Qaisar, and S.~A.
  Alkahtani, ``Can difficulty standing from a chair signal early risk of
  chronic illness? {Insights} from a multi-wave {European} cohort,''
  \emph{Calcified Tissue International}, vol. 117, no.~1, p.~50, 2026.

\bibitem{thalman2020review}
C.~Thalman and P.~Artemiadis, ``A review of soft wearable robots that provide
  active assistance: Trends, common actuation methods, fabrication, and
  applications,'' \emph{Wearable Technologies}, vol.~1, p.~e3, 2020.

\bibitem{kim2019reducing}
J.~Kim, G.~Lee, R.~Heimgartner, D.~Arumukhom~Revi, N.~Karavas, D.~Nathanson,
  I.~Galiana, A.~Eckert-Erdheim, P.~Murphy, D.~Perry \emph{et~al.}, ``Reducing
  the metabolic rate of walking and running with a versatile, portable
  exosuit,'' \emph{Science}, vol. 365, no. 6454, pp. 668--672, 2019.

\bibitem{collins2015reducing}
S.~H. Collins, M.~B. Wiggin, and G.~S. Sawicki, ``Reducing the energy cost of
  human walking using an unpowered exoskeleton,'' \emph{Nature}, vol. 522, no.
  7555, pp. 212--215, 2015.

\bibitem{shveda2022wearable}
R.~A. Shveda, A.~Rajappan, T.~F. Yap, Z.~Liu, M.~D. Bell, B.~Jumet, V.~Sanchez,
  and D.~J. Preston, ``A wearable textile-based pneumatic energy harvesting
  system for assistive robotics,'' \emph{Science Advances}, vol.~8, no.~34, p.
  eabo2418, 2022.

\bibitem{tiboni2022sensors}
M.~Tiboni, A.~Borboni, F.~V{\'e}rit{\'e}, C.~Bregoli, and C.~Amici, ``Sensors
  and actuation technologies in exoskeletons: A review,'' \emph{Sensors},
  vol.~22, no.~3, p. 884, 2022.

\bibitem{nuckols2021individualization}
R.~W. Nuckols, S.~Lee, K.~Swaminathan, D.~Orzel, R.~D. Howe, and C.~J. Walsh,
  ``Individualization of exosuit assistance based on measured muscle dynamics
  during versatile walking,'' \emph{Science Robotics}, vol.~6, no.~60, p.
  eabj1362, 2021.

\bibitem{missiroli2020relationship}
F.~Missiroli, N.~Lotti, M.~Xiloyannis, L.~H. Sloot, R.~Riener, and L.~Masia,
  ``Relationship between muscular activity and assistance magnitude for a
  myoelectric model based controlled exosuit,'' \emph{Frontiers in Robotics and
  AI}, vol.~7, p. 595844, 2020.

\bibitem{mohebbi2020human}
A.~Mohebbi, ``Human-robot interaction in rehabilitation and assistance: a
  review,'' \emph{Current Robotics Reports}, vol.~1, no.~3, pp. 131--144, 2020.

\bibitem{torricelli2020benchmarking}
D.~Torricelli, C.~Rodriguez-Guerrero, J.~F. Veneman, S.~Crea, K.~Briem,
  B.~Lenggenhager, and P.~Beckerle, ``Benchmarking wearable robots: challenges
  and recommendations from functional, user experience, and methodological
  perspectives,'' \emph{Frontiers in Robotics and AI}, vol.~7, p. 561774, 2020.

\bibitem{tucker2015control}
M.~R. Tucker, J.~Olivier, A.~Pagel, H.~Bleuler, M.~Bouri, O.~Lambercy, J.~d.~R.
  Millan, R.~Riener, H.~Vallery, and R.~Gassert, ``Control strategies for
  active lower extremity prosthetics and orthotics: a review,'' \emph{Journal
  of NeuroEngineering and Rehabilitation}, vol.~12, no.~1, p.~1, 2015.

\bibitem{asbeck2015biologically}
A.~T. Asbeck, S.~M. De~Rossi, K.~G. Holt, and C.~J. Walsh, ``A biologically
  inspired soft exosuit for walking assistance,'' \emph{The International
  Journal of Robotics Research}, vol.~34, no.~6, pp. 744--762, 2015.

\bibitem{ma2019real}
X.~Ma, X.~Long, Z.~Yan, C.~Wang, Z.~Guo, and X.~Wu, ``Real-time active control
  of a lower limb exoskeleton based on {sEMG},'' in \emph{Proc. IEEE/ASME Int.
  Conf. Advanced Intelligent Mechatronics (AIM)}.\hskip 1em plus 0.5em minus
  0.4em\relax IEEE, 2019, pp. 589--594.

\bibitem{gordleeva2020real}
S.~Y. Gordleeva, S.~A. Lobov, N.~A. Grigorev, A.~O. Savosenkov, M.~O. Shamshin,
  M.~V. Lukoyanov, M.~A. Khoruzhko, and V.~B. Kazantsev, ``Real-time
  {EEG}--{EMG} human--machine interface-based control system for a lower-limb
  exoskeleton,'' \emph{IEEE Access}, vol.~8, pp. 84\,070--84\,081, 2020.

\bibitem{chi2010dry}
Y.~M. Chi, T.-P. Jung, and G.~Cauwenberghs, ``Dry-contact and noncontact
  biopotential electrodes: Methodological review,'' \emph{IEEE Reviews in
  Biomedical Engineering}, vol.~3, pp. 106--119, 2010.

\bibitem{de2010filtering}
C.~J. De~Luca, L.~D. Gilmore, M.~Kuznetsov, and S.~H. Roy, ``Filtering the
  surface {EMG} signal: Movement artifact and baseline noise contamination,''
  \emph{Journal of Biomechanics}, vol.~43, no.~8, pp. 1573--1579, 2010.

\bibitem{merletti2004surface}
R.~Merletti and P.~A. Parker, Eds., \emph{Electromyography: Physiology,
  Engineering, and Noninvasive Applications}.\hskip 1em plus 0.5em minus
  0.4em\relax Hoboken, NJ, USA: Wiley-IEEE Press, 2004.

\bibitem{beck2008electrode}
T.~W. Beck, T.~J. Housh, J.~T. Cramer, and J.~P. Weir, ``The effects of
  electrode placement and innervation zone location on the electromyographic
  amplitude and mean power frequency versus isometric torque relationships for
  the vastus lateralis muscle,'' \emph{Journal of Electromyography and
  Kinesiology}, vol.~18, no.~2, pp. 317--328, 2008.

\bibitem{hermens2000seniam}
H.~J. Hermens, B.~Freriks, C.~Disselhorst-Klug, and G.~Rau, ``Development of
  recommendations for {SEMG} sensors and sensor placement procedures,''
  \emph{Journal of Electromyography and Kinesiology}, vol.~10, no.~5, pp.
  361--374, 2000.

\bibitem{sul2025emg}
J.-H. Sul, L.~Piyathilaka, D.~Moratuwage, S.~D. Arachchige, A.~Jayawardena,
  G.~Kahandawa, and D.~M.~G. Preethichandra, ``Electromyography signal
  acquisition, filtering, and data analysis for exoskeleton development,''
  \emph{Sensors}, vol.~25, no.~13, p. 4004, 2025.

\bibitem{Toro-Ossaba2025Myoelectric}
A.~Toro-Ossaba, J.~C. Tejada, and D.~Sanín-Villa, ``Myoelectric control in
  rehabilitative and assistive soft exoskeletons: A comprehensive review of
  trends, challenges, and integration with soft robotic devices,''
  \emph{Biomimetics}, vol.~10, no.~4, p. 214, 2025.

\bibitem{islam2013mechanomyography}
M.~A. Islam, K.~Sundaraj, R.~B. Ahmad, N.~U. Ahamed, and M.~A. Ali,
  ``Mechanomyography sensor development, related signal processing, and
  applications: a systematic review,'' \emph{IEEE Sensors Journal}, vol.~13,
  no.~7, pp. 2499--2516, 2013.

\bibitem{smith2017dynamic}
C.~M. Smith, T.~J. Housh, E.~C. Hill, G.~O. Johnson, and R.~J. Schmidt,
  ``Dynamic versus isometric electromechanical delay in non-fatigued and
  fatigued muscle: A combined electromyographic, mechanomyographic, and force
  approach,'' \emph{Journal of Electromyography and Kinesiology}, vol.~33, pp.
  34--38, 2017.

\bibitem{orizio1993muscle}
C.~Orizio, ``Muscle sound: bases for the introduction of a mechanomyographic
  signal in muscle studies,'' \emph{Critical Reviews in Biomedical
  Engineering}, vol.~21, no.~3, pp. 201--243, 1993.

\bibitem{caulcrick2021human}
C.~Caulcrick, W.~Huo, W.~Hoult, and R.~Vaidyanathan, ``Human joint torque
  modelling with {MMG} and {EMG} during lower limb human--exoskeleton
  interaction,'' \emph{IEEE Robotics and Automation Letters}, vol.~6, no.~4,
  pp. 7185--7192, 2021.

\bibitem{jin2026human}
X.~Jin, G.~Gao, W.~Wang, R.~Vaidyanathan, P.~Childs, and Z.~Yu,
  ``Human-in-the-loop capacitive microphone sensors-based muscle sensing system
  for predictive and adaptive exoskeleton assistance,'' \emph{IEEE Robotics and
  Automation Letters}, vol.~11, no.~4, pp. 4657--4664, 2026.

\end{thebibliography}

\end{document}